\documentclass[sigconf]{acmart}

\AtBeginDocument{%
  }

\setcopyright{acmlicensed}
\copyrightyear{2024}
\acmYear{2024}
\acmDOI{XXXXXXX.XXXXXXX}

\acmConference[KDD 2024 Workshop]{Causal Inference and
Machine Learning in Practice.}{August 25--29, 2024}{Barcelona, Spain}
\acmISBN{978-1-4503-XXXX-X/18/06}

\usepackage{listings}
\begin{document}

\title{Using LLMs for Explaining Sets of Counterfactual Examples to Final Users}


\author{Arturo Fredes}
\affiliation{%
  \institution{Departament de Matemàtica i Informàtica, \\ }
  \city{Universitat de Barcelona}
  \country{Spain}}
\email{afredeca7@alumnes.ub.edu}

\author{Jordi Vitrià}
\affiliation{%
  \institution{Departament de Matemàtica i Informàtica, \\ }
  \city{Universitat de Barcelona}
  \country{Spain}}
\email{jordi.vitria@ub.edu}


\begin{abstract}
Causality is vital for understanding true cause-and-effect relationships between variables within predictive models, rather than relying on mere correlations, making them highly relevant in the field of Explainable AI. In an automated decision-making scenario, we can analyze the underlying data-generation process using causal inference methods, which enable us to explain a model’s decision by manipulating features and creating counterfactual examples. These counterfactuals explore hypothetical scenarios where a minimal number of factors are altered, providing end-users with valuable information on how to change their situation.
However, interpreting a set of multiple counterfactuals can be challenging for end-users who are not used to analyze raw data records. In our work, we propose a novel multiple-step pipeline that uses counterfactuals and generates natural language 
explanations of actions that will lead to a change of outcome in classifiers of tabular data using LLMs. This pipeline was thought to guide the LLM through smaller tasks that mimic human reasoning when explaining a decision based on counterfactual cases. We conducted various experiments using a public dataset and proposed a method of closed-loop evaluation to assess the coherence of the final explanation with the counterfactuals as well as the quality of the content. Results show to be promising, although further experiments with other data sets as well as human evaluation should be carried out.

The code produced for this work can be found in the following repository: \url{https://github.com/arturofredes/LLM-4-CFs-Explanation}
\end{abstract}



\keywords{Causality, Explainability, Contrastive explanations, Counterfactuals, LLMs, Transparency}

\received{20 February 2007}
\received[revised]{12 March 2009}
\received[accepted]{5 June 2009}

\maketitle
\begin{table*}[h]
    \caption{Original user data and counterfactual examples created using the Diverse Counterfactual Explanations (DiCE) library \cite{mothilal2020dice}}
    \begin{tabular}{l|rllllllr}
    \hline
       &age & workclass     & education   & status   & occupation    & race   & gender   &   hpw  \\
    \hline
    Original &41 & Private & School & Married & Blue-Collar & White & Male & 30 \\

    \hline
    &41 & Private & Prof-school & Married & Blue-Collar & White & Male & 16  \\
    &41 & Private & School & Divorced & Professional & White & Male & 30  \\
     Counterfactuals &41 & Private & Bachelors & Married & Blue-Collar & White & Male & 19   \\
    &41 & Self-Employed & School & Married & Blue-Collar & White & Female & 30   \\
    &41 & Private & Doctorate & Married & Blue-Collar & White & Male & 30  \\
    \hline
    \end{tabular}
    \label{tab:negative}
\end{table*}
\section{Introduction}
\begin{figure}[h]
  \centering
  \includegraphics[width=0.9\linewidth]{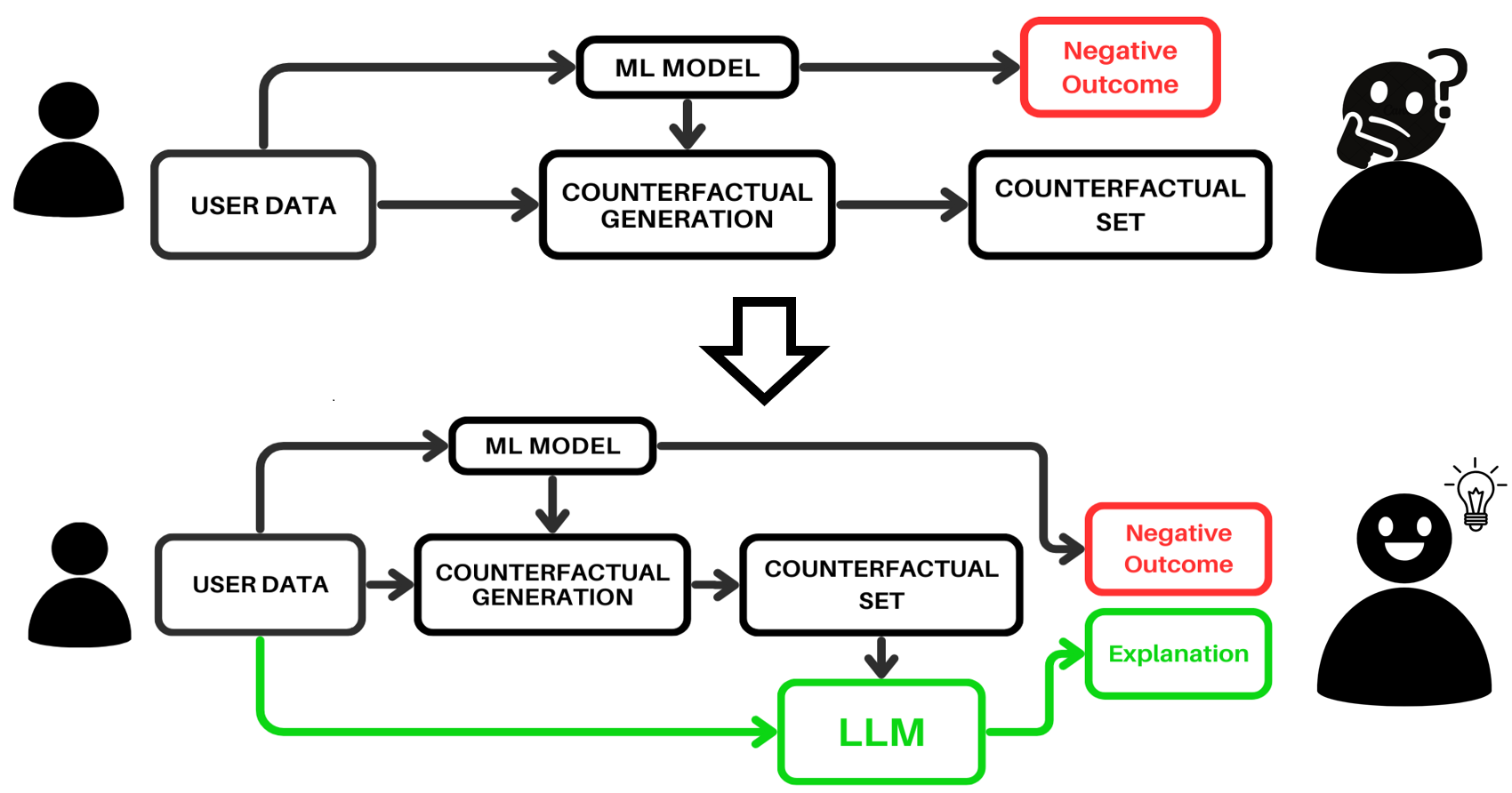}
\caption{Using LLMs to generate a natural language explanation from a set of counterfactuals, which will be more easily interpreted by end users}
  \label{fig:scheme}
\end{figure}
Machine learning systems are becoming increasingly influential in society, especially in critical decision-making areas such as healthcare, education, and justice. However, many of these models function as "black boxes"  \cite{10.1145/3236009} , where inputs are transformed into outputs without transparency into the decision-making process, leading to issues with trust and interpretability for users. This gave birth to the research field of XAI (explainable Artificial Intelligence), where multiple techniques for extracting explanations for model predictions have been proposed. In this work we focus on contrastive explanations through sets of multiple counterfactual examples \cite{molnar2019interpretable}, which are variants of user data with minimal changes in order to flip the prediction. 
 
 When a user is presented with a single counterfactual, extracting conclusions from it is straightforward. Nevertheless, this may not reflect the whole scope of possible actions the user can take. On the other hand, as we increase the number of counterfactuals, drawing conclusions from them becomes difficult for people who are not trained on a data analytic thinking \cite{10.1145/3581641.3584082}. The objective of this work is to evaluate the use of LLMs in generating clear, plain-language explanations for end users without the need for individual case analysis. We guide the LLM through the explanation generation task by breaking it into smaller processes, mimicking the steps a human would follow when explaining an event.


\section{Implementation}
Our work aims to bridge the gap between the ML model's decision and the end user understanding and accepting or contesting that decision. To do that, we would like to generate a clear and actionable explanation for a user who receives a negative outcome, like being denied a loan, using LLMs to automatize the process. Due to the limitations of these models when tackling problems that require planning and multiple step reasoning \cite{mccoy2023embers,stechly2024selfverification}, we broke down the problem into smaller ones, mimicking human reasoning \cite{MILLER20191}. We hope that by doing so, the LLM will be capable of correctly identifying the causes that led to a change of class and synthesise them into a clear explanation which gives advice to the user on how to change to their desired output. In Figure \ref{fig:scheme} we can see a scheme of the different steps taken to generate the final text.

\begin{figure}[h]
  \centering
  \includegraphics[width=0.9\linewidth]{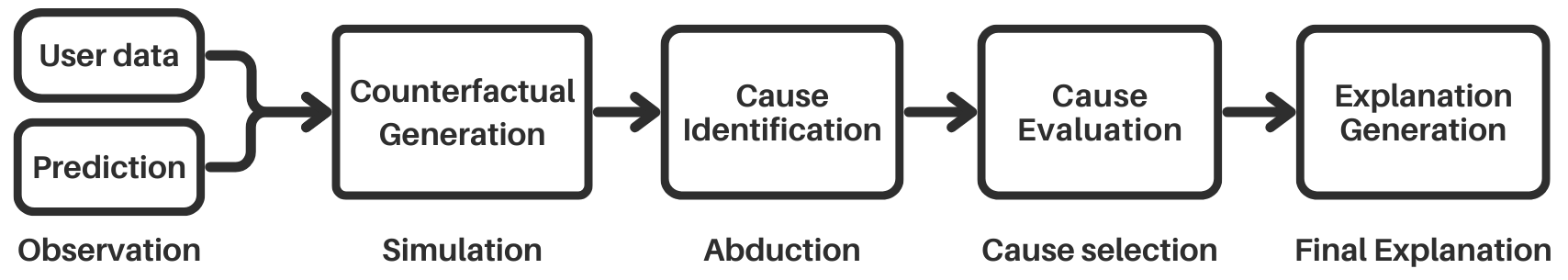}
\caption{Scheme of the different steps taken to generate the final explanation}
  \label{fig:scheme}
\end{figure}

In this section we will explain each of the different steps in more detail, and will use an example of the Adult Dataset \cite{misc_adult_2}  which includes different
features about the socioeconomic status of a set of individuals, and a label indicating whether an individual will earn more than 50k\$ a year.
\subsection{Counterfactual Generation}
One of the most natural approaches to answering a "why" question is to use a contrastive explanation. Humans generate a contrastive explanation by comparing the actual outcome with a hypothetical alternative scenario. The process begins with understanding the event or decision that occurred within a causal context. Then, we consider what could have happened differently by imagining a scenario where a key cause is altered. Finally, we focus on the specific factors or causes that lead to a different outcome in the alternative scenario.

We will simulate alternative scenarios by generating counterfactuals that will flip the decision of the classifier. In our case, we used the DiCE framework \cite{mothilal2020dice}, which is among the most popular, however, other frameworks could be used. DiCE utilizes user data and the ML model to generate counterfactuals through perturbations, approaching the problem as an optimization task. The optimization focuses on maximizing validity (examples that effectively change class), proximity (examples that are as close as possible to the original), and diversity (examples with different causes changing to offer various routes). Additionally, users can restrict the features to be changed in order to avoid some unfeasible counterfactuals (like changing race), and  post-hoc processes are applied to enhance sparsity (changing the minimum number of features possible). By doing this, the obtained set of counterfactuals can provide the user with diverse, feasible paths to the desired outcome, provided they are understood properly. In Table \ref{tab:negative} an example of an individual of the adult dataset predicted to earn less than 50k \$ and five counterfactual examples predicted to earn above that threshold are shown.

\subsection{Extracting causes}
After generating the counterfactual examples, our goal is to identify the primary causes deduced from them. To achieve this, we fed the LLM the set of counterfactual examples as well as the original user data, and asked it to return a list of the main reasons why the user was classified in one way and not the other. This process can be viewed as employing abductive reasoning to generate possible causes. In Table \ref{fig:rules} we can see the prompt used and the output produced by the LLM with the main causes that led to a change of outcome for the example of Table \ref{tab:negative}.

\begin{table}
\raggedright
\caption{Prompt used to extract main causes of the shift in classification and LLM response.}
\begin{tabular}{p{0.95\linewidth}}
\\
\hline
\end{tabular}
\textbf{User Prompt:} I'm providing a negative outcome from a ML-system that predicts whether a person will earn more than 50k \$ a year and your task is to extract the most important observed rules based on a set of counterfactual cases. \\
----- User Data Negative outcome ----- \\
<{\it We insert here a Python dataframe containing original user data}> \\

----- Positive counterfactual outcome ----- \\
<{\it We insert here a Python dataframe containing the generated counterfactual examples}> \\

----- Rules ----- \\
<List of Rules> \bigskip

\begin{tabular}{p{0.95\linewidth}}
\\
\hline
\end{tabular}
\textbf{GPT-4o answer:}
Based on the provided negative assessment outcome and the positive counterfactual outcomes, here are the most important observed rules:

\begin{enumerate}
    \item \textbf{Higher Education (Prof-school, Bachelors, Doctorate) Leads to Higher Income}:
    \begin{itemize}
        \item The individual in the negative outcome has a "School" level of education.
        \item In the positive counterfactuals, individuals with "Prof-school", "Bachelors", and "Doctorate" education levels are shown to have a higher income.
    \end{itemize}
    
    \item \textbf{Marital Status Being Divorced Can Lead to Higher Income}:
    \begin{itemize}
        \item In the counterfactual case where the individual is "Divorced", the income is higher even with the same "School" level of education and same occupation.
    \end{itemize}
    
    \item \textbf{Type of Workclass Can Influence Income}:
    \begin{itemize}
        \item The individual in the negative outcome is in the "Private" workclass.
        \item One of the counterfactuals shows that being "Self-Employed" with the same education level leads to a higher income.
    \end{itemize}
    
    \item \textbf{Occupation Being Professional Leads to Higher Income}:
    \begin{itemize}
        \item The individual in the negative outcome is in a "Blue-Collar" occupation.
        \item In the counterfactuals, an occupation such as "Professional" leads to a higher income, even with the same education level and marital status.
    \end{itemize}
    
    \item \textbf{Hours Per Week Less Influential in this Case}:
    \begin{itemize}
        \item Despite variations in hours per week (16, 19, 30) in the positive counterfactuals, the individuals still have higher incomes, indicating that hours per week alone are less influential compared to education level and occupation.
    \end{itemize}
\end{enumerate}

These rules highlight key factors that can influence the income prediction in this context.

\begin{tabular}{p{0.95\linewidth}}
\\
\hline
\end{tabular}
\bigskip
\bigskip
\bigskip

\label{fig:rules}
\end{table}

\subsection{Cause Evaluation and Selection}

\begin{table}
\raggedright
\caption{Prompt used to generate code that counts how many counterfactuals contain each cause, code produced by the LLM and results of code execution.}

\begin{tabular}{p{0.95\linewidth}}
\hline
\end{tabular}
\textbf{User Prompt:} I'm providing a negative outcome from a {ML-system}, a set of counterfactual cases that flip the decision of the system and the main rules inferred from the counterfactuals.
You should generate python code to count how many of the counterfactuals are consistent with the rule. The code should create a df with the counterfactuals provided and then check for each rule how many of them follow the rules. Order the rules. Finally, you should print the results.\\

----- User Data Negative outcome ----- \\
<{\it We insert here a Python dataframe containing original user data}> \\

----- Positive counterfactual outcome ----- \\
<{\it We insert here a Python dataframe containing the generated counterfactual examples}> \\

----- Rules ----- \\
<{\it We insert here a list with the generated rules}>  \\

----- Dataset info -----

The following info about the dataset is available: \\
<{\it We insert here public information about the dataset}> \\
----- Code -----\\
```\\
{\tt import pandas as pd}\\
{\tt \# complete this code}\\
```\bigskip

\begin{tabular}{p{0.95\linewidth}}
\\
\hline
\end{tabular}
\textbf{GPT-4o answer:}
\begin{lstlisting}[basicstyle=\tiny, breaklines = true]
import pandas as pd

# Original negative outcome
original = pd.DataFrame({'age': [41], 'workclass': ['Private'], 'education': ['School'], 'marital_status': ['Married'], 'occupation': ['Blue-Collar'], 'race': ['White'], 'gender': ['Male'], 'hours_per_week': [30], 'income': [0]})

# Counterfactuals
counterfactuals = pd.DataFrame({
    'age': [41, 41, 41, 41, 41],
    'workclass': ['Private', 'Private', 'Private', 'Self-Employed', 'Private'],
    'education': ['Prof-school', 'School', 'Bachelors', 'School', 'Doctorate'],
    'marital_status': ['Married', 'Divorced', 'Married', 'Married', 'Married'],
    'occupation': ['Blue-Collar', 'Professional', 'Blue-Collar', 'Blue-Collar', 'Blue-Collar'],
    'race': ['White', 'White', 'White', 'White', 'White'],
    'gender': ['Male', 'Male', 'Male', 'Female', 'Male'],
    'hours_per_week': [16, 30, 19, 30, 30],
    'income': [1, 1, 1, 1, 1]})

# Rule 1: Higher education (Prof-school, Bachelors, Doctorate)
rule1 = counterfactuals['education'].isin(['Prof-school', 'Bachelors', 'Doctorate']).sum()
# Rule 2: Marital status being Divorced...
rule2 = (counterfactuals['marital_status'] == 'Divorced').sum()
# Rule 3: Type of workclass can influence income (Self-Employed...
rule3 = (counterfactuals['workclass'] == 'Self-Employed').sum()
# Rule 4: Occupation being Professional leads to higher income
rule4 = (counterfactuals['occupation'] == 'Professional').sum()
# Rule 5: Hours per week (not directly counted as a rule)

# Print results
print("Number ... (Higher education):", rule1)
print("Number ... (Marital status being Divorced):", rule2)
print("Number ... (Self-Employed workclass):", rule3)
print("Number ... (Occupation being Professional):", rule4)
    
\end{lstlisting}

\begin{tabular}{p{0.95\linewidth}}
\\
\hline
\end{tabular}
\textbf{RESULT OF CODE EXECUTION:}

Number of counterfactuals following Rule 1 (Higher education): 3 \\
Number of counterfactuals following Rule 2 (Marital status being Divorced): 1 \\
Number of counterfactuals following Rule 3 (Self-Employed workclass): 1 \\
Number of counterfactuals following Rule 4 (Occupation being Professional): 1

\begin{tabular}{p{0.95\linewidth}}
\\
\hline
\end{tabular}
\label{fig:rulecheck}
\end{table}
Once a list is generated, we need to verify its correctness and identify the most relevant causes to the explanation. To achieve this, we use a program-aided prompt  \cite{gao2023pal} to instruct the LLM to generate Python code that counts how many times a cause was responsible (or partially responsible) of flipping the outcome by counting the number of counterfactual examples that contained the proposed change. By generating and executing this code, we can externally confirm the LLM's accuracy and rank each of the main reasons in the generated list. Although a cause appearing in all counterfactuals does not necessarily imply it is essential, it indicates higher relevance. 

In order for the LLM to generate this code we feed it the original user data, the set of counterfactual examples, the output with the main causes of the previous step and some information about the columns and possible values of the dataset in order to make the code generation more reliable. This can be seen in Table \ref{fig:rulecheck} as well as the code generated by the LLM and the results of the execution of the generated code.

\subsection{Generating explanation}

Finally, we leverage all the information that has been produced and feed it to the LLM. We ask it to generate a final explanation in plain language that emphasizes actions that can be taken by the user in order to change to the desired class. This is were the LLM is most useful, since we assume that in its training it will have captured the correct form of human interaction, and that by crafting the correct prompt and using the previous information, the explanation will satisfy the final user.

In Table \ref{fig:explanation} the prompt used and the final textual explantion generated by the LLM are shown.

\begin{table*}
\raggedright
\caption{Prompt used to generate an explanation of the counterfactuals and LLM response}

\begin{tabular}{p{0.95\linewidth}}
\\
\hline
\end{tabular}
\textbf{User prompt:} A person has been classified in the negative class of {ML-system}. The data is the following.\\

----- User Data Negative outcome ----- \\
<{\it We insert here a Python dataframe containing original user data}> \\

----- Positive counterfactual outcome ----- \\
<{\it We insert here a Python dataframe containing the generated counterfactual examples}> \\

----- Rules ----- \\
By generating counterfactual, we obtained the following rules:\\
<{\it We insert here a list with the generated rules}>  \\

----- Results -----\\
We have checked that the rules are followed by $n$ counterfactual:\\
<{\it We insert here code execution results}>  \\

----- Dataset info -----\\
The following info about the dataset is available:\\
<{\it We insert here public information about the dataset}> \\
----- Explanation -----\\
Given this information, provide an explanation to the user in plain language so that he/she can improve their chances of changing class. It should be as clear as possible and call to action. Consider that the higher amount of counterfactual that follow the rule, the more important that rule is. \\
```{\tt <explanation>}
```\bigskip

\begin{tabular}{p{0.95\linewidth}}
\\
\hline
\end{tabular}
\textbf{GPT-4o answer:} Based on the analysis of your current situation and the observed patterns in the data, here are some clear steps you can take to improve your chances of earning more than \$50K a year:

\paragraph*{1. Pursue Higher Education}
The most influential factor is education. Individuals with higher education levels such as Professional School, Bachelors, and Doctorate degrees are more likely to earn a higher income. Currently, you have a "School" level of education. Investing in further education could significantly improve your economic prospects. Consider enrolling in higher education programs or professional courses to enhance your qualifications.

\paragraph*{2. Explore Different Occupations}
Another key factor is the type of occupation. Individuals in "Professional" roles tend to earn higher incomes compared to those in "Blue-Collar" jobs. If possible, seek opportunities or training that could transition you into a more professional role within your field or a new industry that values your skills.

\paragraph*{3. Consider Marital Status Impact}
While this factor is less controllable, data shows that individuals who are "Divorced" have higher incomes even with the same education and occupation. This might reflect a variety of socio-economic factors. While changing your marital status is not a practical recommendation, it's important to be aware of how life circumstances can impact income.

\paragraph*{4. Evaluate Workclass Opportunities}
Being "Self-Employed" is another observed factor that can lead to higher income compared to being in the "Private" workclass. If you have entrepreneurial skills or a business idea, exploring self-employment might be a viable path to increasing your income.

\paragraph*{5. Hours Per Week is Less Critical}
Interestingly, the total number of hours you work per week seems to be less influential compared to education and occupation. While working more hours can help, focusing on improving your education and transitioning to a higher-income occupation will likely have a greater impact on your income level.

\paragraph*{Action Steps}
\begin{enumerate}
    \item \textbf{Education}: Research higher education programs or professional courses that align with your career goals.
    \item \textbf{...}
\end{enumerate}

\begin{tabular}{p{0.95\linewidth}}
\\
\hline
\end{tabular}
\label{fig:explanation}
\end{table*}
\section{Evaluation}
\subsection{What makes a good explanation?}

Before delving into how to evaluate the process of the LLM, it is important to introduce some concepts on what makes a good explanation from the social sciences perspective \cite{MILLER20191}, which will help us create a better evaluation.

\paragraph{Contrastive Explanations:}
People are usually not just interested in why something happened, but why it happened instead of something else. Counterfactual examples are contrastive by nature, which makes them a perfect starting point for building an explanation.
\paragraph{Biased Selection in Explanations:}
People don't usually expect to be presented with the whole causal chain that leads to an event happening. We often choose only one or two possible reasons out of many when explaining something, and these choices can be biased. This selective process is influenced by cognitive biases, meaning that we might prefer certain explanations over others, not because they are more accurate, but because they align better with our way of thinking. Some of these biases are selecting simpler and more general causes, selecting more recent and abnormal events as causes, or using necessary causes over sufficient ones. The latter is especially relevant in our work, because if a cause appears in all counterfactuals  it has more chances of being necessary and therefore we should pay attention to it.
\paragraph{Social Dynamics of Explanations:}
Explanations are social, functioning as a knowledge transfer mechanism within interactions or conversations. A person who receives an explanation will not be satisfied if it is not aligned with his or hers prior beliefs and if it is not given in the correct form. We assume that the LLM has captured this common knowledge in its training and that it will be capable of writing the explanation in a correct manner given the right instructions.

\subsection{Closed Loop Evaluation}

As it is the case for most of LLM applications, evaluating how "good" these explanations are is not an easy task. It would be ideal to have a group of people manually analyzing the outputs of the LLM, but this very time consuming and expensive. Because of this, we decided to make a "closed loop" evaluation of LLM's understanding of the counterfactuals and causes of the change. We did so by asking the LLM to generate a counterfactual example \cite{gat2023faithful} at the end of the reasoning, using only the original case, information of the dataset and the final explanation. If the LLM is capable of generating a counterfactual example from the explanation it means that the important information was correctly synthesized in it. Additionally, we checked some metrics to evaluate the quality of the explanation. To obtain all of these metrics, a final prompt was used to create a table with all the information and save it in a csv file. After extracting and executing the code generated by the LLM we can apply external operations on the table \cite{wang2024chainoftable}. The different metrics evaluated are the following:

\begin{itemize}
    \item {\bf Validity:} The final example produced by the LLM must belong to the desired class. This will indicate that the explanation contained the necessary information in order to produce the flip in classification.
    \item {\bf Consistency with causes:}
The example we generate must be consistent with the identified causes that were reasoned in the previous steps, so we will check how many of them it follows. As we mentioned, people prefer simpler explanations and only the main causes, so the ratio of causes used in the explanation and the use of the top ranked causes could be seen as a proxy of explanation quality.
\item {\bf Novelty:}
Since the dataset we are using is quite well known and available on the internet, we should check that the LLM is not taking samples it has memorized from the original dataset.
\item {\bf Feasibility:}
We would also like to check if the example given could be applied in the real world. Since the perturbation method can change the selected variables freely, it can produce some examples that may not be applicable in the real world, for example, suggesting to change to a job for which we are not qualified without changing our education. This step is optional and can only be done if one has access to the causal graph \cite{Neal2020CausalInferenceBook}. It will be left as future work.
\end{itemize}

In Appendix A we show the prompts and the pipeline used for this evaluation.

\section{Experiments and Results}

As in the illustrative example used to explain the method, in our experiments we worked with the "Adult" dataset \cite{misc_adult_2}. A Random Forest classifier was trained, to predict whether an individual will earn more than 50k\$ a year and we selected 100 cases in which the system predicted that an individual would earn less than 50k\$ per year. Explanations were generated with our proposed pipeline using GPT-4o \cite{openai2024gpt4o} as the LLM and DiCE allowing all features to vary freely to generate the counterfactuals. We evaluated the explanations with the method method mentioned in the previous section. We varied the size of the sets of counterfactuals between 1, 3 and 5 to observe the effects of adding more counterfactuals to the causes identified, and how the LLM selected them. Furthermore, we started by trying two different prompting techniques, Zero-Shot and Few-Shot \cite{brown2020language}, to see if there were improvements in the results.

Since generating the counterfactuals is an optimization problem with random initial conditions, the generated counterfactuals can be different each time the algorithm is called. On top of this, the responses of the LLM are also stochastic \cite{vaswani2023attention}, and vary in each generation. We used Tree of Thought (ToT) \cite{yao2023tree} to take advantage of this and try to create more diversity in the explanations. We used both of the prompting techniques previously mentioned and created three different explanations and sets  of identified causes of the change starting from different sets of 5 counterfactuals. We then leveraged all the information to create a final explanation and hoped that this would add diversity to the causes and lead to a better final explanation. In Appendix B the prompts for this method are shown.

The results of the experiments are shown in Table \ref{tab:experiments}. 

\begin{table*}[h]
  \caption{Results of the experiments. Explanations for 100 cases of the adult dataset were generated and evaluated using different amounts of counterfactuals (CFs) and three different prompting techniques, Zero-Shot, One-Shot and Tree of Thought (ToT).}
  \label{tab:freq}
  \begin{tabular}{l|ccc|ccc|c}
    \toprule
    && One Shot&&&Zero Shot&&ToT \\
    \midrule
    Mean Values&1 CF & 3 CFs &5 CFs&1 CF & 3 CFs &5 CFs & 5 CFs\\
    \midrule
    Validity & 75\% & 73\% & 79\%& 86\% & 84\% & 93\%&72\%\\
    Causes Identified & 2.98 & 3.18 & 3.71& 2.02 & 3.18 & 4.31&9.61\\
     Causes used & 78\% & 73\% & 77\%& 85\% & 64\% & 75\%& 44\%\\
    1st Cause used & 91\% & 82\% & 90\%& 92\% & 74\% & 95\%&70\%\\
    2nd Cause used & 79\% & 78\% & 88\%& 78\% & 61\% & 71\%&55\%\\
    3rd Cause used& 68\% & 64\% & 69\%& 67\% & 53\% & 71\%&55\%\\
    In the data & 14\% & 10\% & 15\%& 22\% & 23\% & 27\%&19\%\\
  \bottomrule
\end{tabular}
\label{tab:experiments}
\end{table*}

Validity was over 70\% in all the experiments we did, showcasing the method proposed is able to generate explanations that would contain the necessary information to generate a counterfactual example that flips the decision of the classifier. Furthermore, we can see for the different prompt types that by using larger sets of counterfactual examples the validity improves. When more examples are given, more reasons for the given classification are found, and therefore a larger amount of different paths to obtain the desired outcome can be observed. This could be a cause of the increase in validity.

The best results in terms of validity were obtained when using a Zero-shot strategy. However, this does not mean that the explanation is better in terms of quality, and a human evaluation should be made to determine this. When using Tree of Thought, larger sets of causes were extracted, which is what we expected, although validity decreased with respect to the other experiments. Again, this does not mean that the quality is worse.

In all the cases, the final counterfactual generated by the LLM is consistent with the deduced causes, which indicates that they are contained in the explanation. Furthermore, we can see that not all the causes are applied to the example, which is what we expect if a good explanation was given. In addition to this, we see that the causes selected are consistent with the rank we made, and the most relevant cause is selected in most of the cases, having the highest percentage of selection followed by the second and then the third.

The percentage of generated examples present in the dataset is below 30\% in all the experiments. This could be coincidences, since the dataset is large and there is not a lot of variability in the features which are usually changed like education level and profession. However, we cannot be certain about this and further experiments should be carried out with other less known data sets.

\section{Conclusions}
In our work we present a method to produce
automatic explanations of a set of counterfactuals using LLMs. We have seen that the LLM is capable of correctly extracting possible causes that will lead to a flip of the decision of a classifier by analyzing these counterfactual examples, select the main reasons according to their relevance and generate a final explanation in natural language that can be interpreted by the final user. We have also observed that increasing the amount of counterfactuals increases the amount of times the LLM is capable of generating a counterfactual of the opposite class using only the natural language explanation, indicating better information given. Furthermore, using larger sets of counterfactuals and prompting techniques like Tree of Thought can lead to more diversity of causes in the final explanation.
Overall, our method  shows to be a promising option to complement sets of counterfactual examples in order to make them more understandable for the final user. However, further experiments should be made with other datasets, and with human evaluators in order to assess the quality if the natural language explanations.

\bibliographystyle{ACM-Reference-Format}
\bibliography{bib}


\appendix
\section{Closed Loop Evaluation Prompts and Pipeline}

In this section we will show the prompts and the different intermediate results obtained in order to better illustrate the process of closed loop evaluation of the natural language explanation. We will start by asking the LLM to generate a final counterfactual by using only the original user data, the textual explanation in order to know which changes it can make, and finally some information about the name of the columns and the possible values they can take, so it creates an example that could belong to the dataset. This example will be created by the LLM with python code, and it will be saved into a temporary csv file. We can see the prompt, and the output of the LLM in Table \ref{fig:examplecode}.

Once the example is created by executing the code externally, we will prompt the LLM in order to compare the example generated from the textual explanation to the previously identified causes and their importance. In this part we will feed the LLM the causes we identified, the results of the code that ranked them according to how many of the counterfactuals contained that cause, original user data and information about the dataset. The result of this process will be a csv with the list of causes indicating their importance and if they are represented in the final counterfactual. In Table \ref{fig:evalcode} an example of the process is given. Finally, we will analyze the table created by the LLM to extract the final metrics like in Table \ref{tab:evalfinal}.


\begin{table*}
\raggedright
\caption{Prompt used to generate a final counterfactual example by using the original data, the explanation and information about the dataset. Code produced by the LLM and results from the execution.}

\begin{tabular}{p{0.95\linewidth}}
\\
\hline
\end{tabular}
\textbf{User prompt:} A person has been classified in the negative class of {ML-system}. The data is the following:\\
----- User Data Negative outcome ----- \\
<{\it We insert here a Python dataframe containing original user data}> \\

----- Explanation -----\\
The following explanation was given in order to try and change the class.\\
<{\it We insert here the provided explanation}> \\

----- Dataset info -----\\
The following info about the dataset is available:\\
<{\it We insert here public information about the dataset}> \\

----- Example -----\\
Given this information, provide an example in the format of a pandas dataframe that would be in the positive class. Complete the code below and note that it is very important to use the name 'temp\_csv.csv', since later processes rely on it.\\

```\\
{\tt import pandas as pd}\\
{\tt df = pd.DataFrame(...) \#complete this line}\\
{\tt df.to\_csv('temp\_csv.csv', index = False)}\\
```\bigskip

\begin{tabular}{p{0.95\linewidth}}
\\
\hline
\end{tabular}

\textbf{GPT-4o answer:}
\begin{lstlisting}[basicstyle=\tiny, breaklines=true]
import pandas as pd

# Define the data for the DataFrame
data = {'age': [41], 'workclass': ['Self-Employed'], 'education': ['Bachelors'], 'marital_status': ['Married'], 'occupation':['Professional'], 'race': ['White'], 'gender': ['Male'], 'hours_per_week':[30], 'income': [1]}

# Create the DataFrame
df = pd.DataFrame(data)

# Save to csv
df.to_csv('temp_csv.csv', index=False)
\end{lstlisting}

\textbf{RESULT OF CODE EXECUTION:}
\bigskip

\begin{tabular}{rllllllr}
    \hline
       age & workclass     & education   & status   & occupation    & race   & gender   &   hpw  \\
    \hline
    41 & Self-Employed	 & Bachelors	 & Married & Professional & White & Male & 30 \\
    \hline
\end{tabular}
\begin{tabular}{p{0.95\linewidth}}
\\
\hline
\end{tabular}
\label{fig:examplecode}
\end{table*}

\begin{table*}
\raggedright
\caption{Prompt used to count how many of the causes are present in the final example and give the results in table format. Code produced by LLM and results.}

\begin{tabular}{p{0.95\linewidth}}
\\
\hline
\end{tabular}
\textbf{User prompt:}
I'm providing a negative outcome from a {ML-system}. A counterfactual example in the format of a single row dataframe was created in temp\_csv from the rules that are also provided. Give some code to check the number of rules followed by the example. The result must be given in the format of a dataframe and saved as a csv. The dataframe must have columns 'Rule' with the text of the rule, 'Importance' with the number of counterfactuals follow each rule, and 'In explanation' (1 or 0) depending if the final example follows the explanation or not. It is very important to save the csv as 'evaluation.csv'.\\
---- User Data Negative outcome ----- \\
<{\it We insert here a Python dataframe containing data from Table 2}> \\

----- Rules ----- \\
By generating counterfactual, we obtained the following rules:\\
<{\it We insert here a list with the generated rules}>  \\

----- Results -----\\
We have checked that the rules are followed by $n$ counterfactual:\\
<{\it We insert here code execution results}>  \\

----- Dataset info -----\\
The following info about the dataset is available:\\
<{\it We insert here public information about the dataset}> \\

----- Code -----\\
```
{\tt import pandas as pd}\\
{\tt df = pd.read\_csv('temp\_csv.csv')}\\

{\tt \# Complete code}\\

{\tt \# Save to csv}\\
{\tt df\_final.to\_csv('evaluation.csv', index = False)}\\
```\bigskip

\begin{tabular}{p{0.95\linewidth}}
\\
\hline
\end{tabular}
\textbf{GPT-4o:}
\begin{lstlisting}[basicstyle=\tiny]
import pandas as pd

# Read example
df = pd.read_csv('temp_csv.csv')

# Define the data for the DataFrame
data = {
    'Rule': [
        'Higher education (Prof-school, Bachelors, Doctorate) leads to higher income.',
        'Marital status being Divorced can lead to higher income.',
        'Type of workclass being Self-Employed can influence income.',
        'Occupation being Professional leads to higher income.',
        'Hours per week less influential in this case.'
    ],
    'Importance': [3, 1, 1, 1, 0],  # Counterfactuals following each rule
    'In explanation': [0, 0, 0, 0, 0]  # Initial values, will update based on the example
}

# Create the DataFrame
df_final = pd.DataFrame(data)

# Check if the example follows each rule
if df['education'].iloc[0] in ['Prof-school', 'Bachelors', 'Doctorate']:
    df_final.at[0, 'In explanation'] = 1
if df['marital_status'].iloc[0] == 'Divorced':
    df_final.at[1, 'In explanation'] = 1
if df['workclass'].iloc[0] == 'Self-Employed':
    df_final.at[2, 'In explanation'] = 1
if df['occupation'].iloc[0] == 'Professional':
    df_final.at[3, 'In explanation'] = 1
if df['hours_per_week'].iloc[0] < 40:  # Assumption based on the provided example
    df_final.at[4, 'In explanation'] = 1

# Save to csv
df_final.to_csv('evaluation.csv', index=False)
\end{lstlisting}

\begin{tabular}{p{0.95\linewidth}}
\\
\midrule
\end{tabular}
\textbf{RESULT OF CODE EXECUTION:}
\bigskip

    \begin{tabular}{lll}
        \hline
        \textbf{Rule} & \textbf{Importance} & \textbf{In explanation} \\
        \hline
        Higher education (Prof-school, Bachelors, Doct... & 3 & 1 \\
        Marital status being Divorced can lead to high... & 1 & 0 \\
        Type of workclass being Self-Employed can infl... & 1 & 1 \\
        Occupation being Professional leads to higher ... & 1 & 1 \\
        Hours per week less influential in this case... & 0 & 1 \\
        \hline
    \end{tabular}

\label{fig:evalcode}
\end{table*}

\begin{table*}[h]
    \centering
    \caption{Final `` Closed loop evaluation" of the explanation.}
    \begin{tabular}{lc}
    \hline
            Metric & Value \\
        \hline
            Validity &  True\\
            Causes Identified  & 5 \\
            Causes used  &  4\\
            1st Cause used & 1 \\
            2nd Cause used &  0\\
            3rd Cause used &  1\\
            In the data &  False\\
        \hline
    \end{tabular}
    \label{tab:evalfinal}
\end{table*}

\section{Tree of Thought Prompts}
In this section we will explain more thoroughly how we used the Tree of Thought strategy to generate explanations more diverse in the causes.
In Table \ref{fig:totexplanation} we show the prompt used to generate the explanation with a Tree of Thought strategy. We generate the explanation and rules three times, twice using a Zero-Shot and once using a One-shot prompt. In the prompt we can see how we leverage the three executions in order to generate a more diverse explanation.
\begin{table*}
\raggedright
\caption{Prompt used to generate an explanation of the counterfactuals and LLM response}
\begin{tabular}{p{0.95\linewidth}}
\toprule
\\
\end{tabular}
\textbf{User prompt:} A negative outcome from a {ML-system} was provided to several systems that explain why that case is negative analyzing counterfactuals, generating rules and evaluating them.
The results of the system are the following.\\
\textbf{System 1}: \\
----- Rules ----- \\
By generating counterfactual, we obtained the following rules:\\
<{\it We insert here a list with the generated rules}>  \\

----- Results -----\\
We have checked that the rules are followed by $n$ counterfactual:\\
<{\it We insert here code execution results}>  \\

----- Explanation -----\\
<{\it Explanation generated by the LLM}> \\
\bigskip

\textbf{System 2}: \\
----- Rules ----- \\
By generating counterfactual, we obtained the following rules:\\
<{\it We insert here a list with the generated rules}>  \\

----- Results -----\\
We have checked that the rules are followed by $n$ counterfactual:\\
<{\it We insert here code execution results}>  \\

----- Explanation -----\\
<{\it Explanation generated by the LLM}> \\

\bigskip

\textbf{System 3}: \\
----- Rules ----- \\
By generating counterfactual, we obtained the following rules:\\
<{\it We insert here a list with the generated rules}>  \\

----- Results -----\\
We have checked that the rules are followed by $n$ counterfactual:\\
<{\it We insert here code execution results}>  \\

----- Explanation -----\\
<{\it Explanation generated by the LLM}> \\
\bigskip

Given this information, provide an explanation to the user in plain language so that he/she can improve their chances of changing class. It should be as clear as possible and call to action. Consider that the higher amount of counterfactuals that follow the rule, the more important that rule is. 

\begin{tabular}{p{0.95\linewidth}}
\\
\hline
\end{tabular}
\label{fig:totexplanation}
\end{table*}
\end{document}